\documentclass[10pt,twocolumn,letterpaper]{article}

\usepackage[pagenumbers]{cvpr} 

\usepackage{arydshln}  
\usepackage{multirow}
\usepackage{graphicx}  

\usepackage{xspace}

\usepackage{amsmath}
\usepackage{amssymb}
\usepackage{graphicx}
\usepackage{booktabs}
\usepackage{algorithm}
\usepackage{algorithmic}

\definecolor{cvprblue}{rgb}{0.21,0.49,0.74}
\usepackage[pagebackref,breaklinks,colorlinks,allcolors=cvprblue]{hyperref}

\def\paperID{13516} 
\def\confName{CVPR}
\def\confYear{2026}
\newcommand{\paper}{Splatent\xspace}
\usepackage{dblfloatfix}

\title{Splatent: Splatting Diffusion Latents for Novel View Synthesis}

\author{
Or Hirschorn$^{\dagger1,2}$,
Omer Sela$^{\dagger1,2}$,
Inbar Huberman-Spiegelglas$^1$,
Netalee Efrat$^1$,\\
Eli Alshan$^1$,
Ianir Ideses$^1$,
Frederic Devernay$^1$,
Yochai Zvik$^1$,
Lior Fritz$^1$\\
\\
$^1$Amazon Prime Video
\quad \quad
$^2$Tel-Aviv University\\
\\
\texttt{\url{https://orhir.github.io/Splatent}}
}

\begin{document}
\twocolumn[{
\maketitle
\renewcommand\twocolumn[1][]{#1}
\begin{center}
    \centering
    \includegraphics[ width=0.85\linewidth]{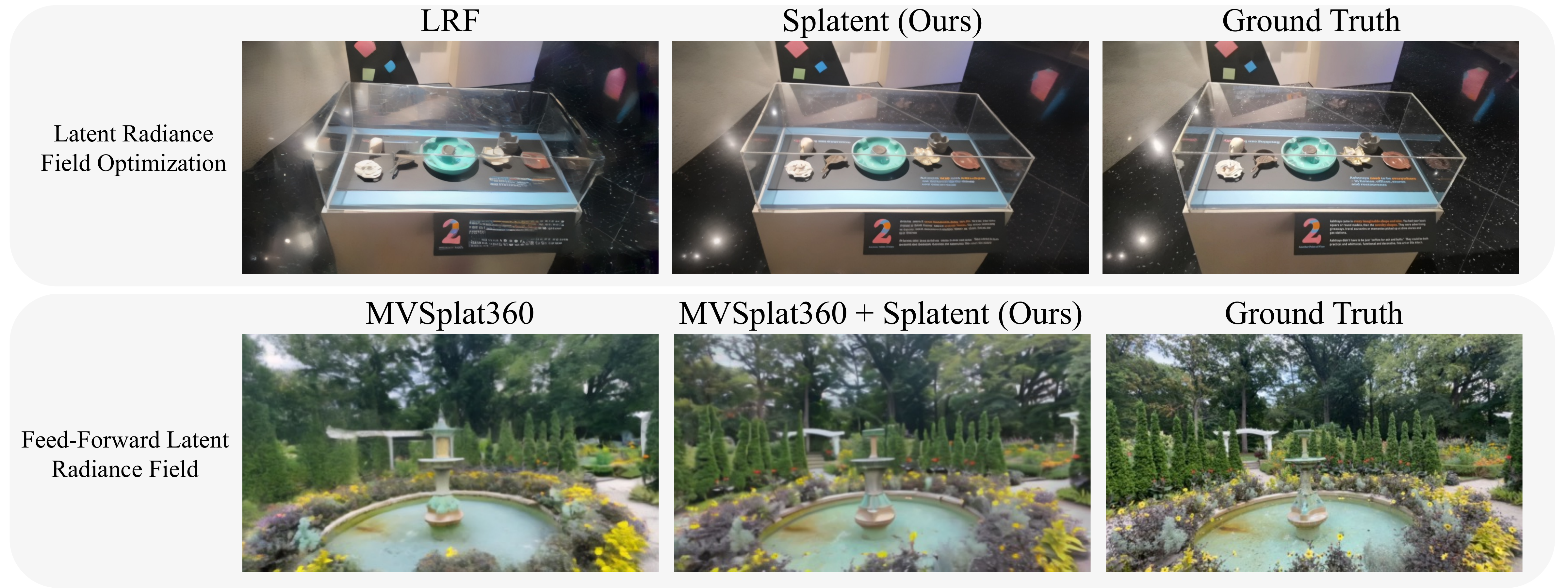}
    \captionof{figure}{\textbf{Novel view synthesis from a latent-space radiance field.} \paper is a principled framework to enhance rendered novel views from a radiance field in the latent space of diffusion VAEs. We demonstrate improvements in image quality in the setting of test-time latent radiance field optimization, compared to LRF~\cite{zhou2025latent}. In addition, we show how \paper can be connected within a latent-based feed-forward model like MVSplat360~\cite{chen2024mvsplat360} to enhance the results and reduce hallucinations.
    }
    \label{fig:teaser}
\end{center}
}]
\def\thefootnote{$\dagger$}\footnotetext{Work was done during an internship at Amazon}

\begin{abstract}

Radiance field representations have recently been explored in the latent space of VAEs that are commonly used by diffusion models. This direction offers efficient rendering and seamless integration with diffusion-based pipelines. 
However, these methods face a fundamental limitation: The VAE latent space lacks multi-view consistency, leading to blurred textures and missing details during 3D reconstruction. 
Existing approaches attempt to address this by fine-tuning the VAE, at the cost of reconstruction quality, or by relying on pre-trained diffusion models to recover fine-grained details, at the risk of some hallucinations. 
We present \paper, a diffusion-based enhancement framework designed to operate on top of 3D Gaussian Splatting (3DGS) in the latent space of VAEs. 
Our key insight departs from the conventional 3D-centric view: rather than reconstructing fine-grained details in 3D space, we recover them in 2D from input views through multi-view attention mechanisms. This approach preserves the reconstruction quality of pretrained VAEs while achieving faithful detail recovery.
Evaluated across multiple benchmarks, \paper establishes a new state-of-the-art for VAE latent radiance field reconstruction. We further demonstrate that integrating our method with existing feed-forward frameworks, consistently improves detail preservation, opening new possibilities for high-quality sparse-view 3D reconstruction.


\end{abstract}

\section{Introduction}

Radiance field representations such as NeRFs~\cite{mildenhall2020nerf} and 3D Gaussian splatting (3DGS)~\cite{kerbl20233dgaussians} have established new standards for photorealistic novel view synthesis. Concurrently, diffusion models have advanced rapidly, with state-of-the-art models typically operating in a compressed latent space obtained through a variational autoencoder (VAE)~\cite{rombach2022high}. Recently, some works explored the idea of radiance field representations directly in this space~\cite{zhou2025latent,aumentado2023reconstructive,chen2024repaint123,metzer2023latent, park2023ed,chen2024mvsplat360}.

Radiance field representations in diffusion latent spaces offer several advantages. Operating at compressed spatial resolutions substantially accelerates both the optimization time and rendering speeds of radiance fields~\cite{metzer2023latent,park2023ed,aumentado2023reconstructive,chen2024repaint123}. Moreover, directly predicting the 3D Gaussian features in latent space enables end-to-end training of feed-forward 3D reconstruction models~\cite{chen2024mvsplat360,schwarz2025generative,go2025vist3a,tang2024lvsm}, as gradients propagate straight to the 3D model without attenuation through an encoder. Prior work showed that predicting RGB values, encoding them, and then feeding them into a generative model yields inferior results~\cite{chen2024mvsplat360}. Operating in latent space avoids this bottleneck and allows diffusion priors to refine the 3D representation directly.
However, directly rendering these latents from a latent radiance field faces obstacles. We demonstrate, consistent with concurrent observations~\cite{zhou2025latent}, that VAE latent spaces used in modern latent diffusion models encode high-frequency details in fundamentally \emph{view-inconsistent} ways~\cite{zhou2025latent,schwarz2025generative}. 
These inconsistencies can severely degrade 3D reconstruction, resulting in blurred textures and missing fine details (see Fig.~\ref{fig:spectral_all_channels}). 
Recent attempts to address this either compromise decoder quality, as in VAE fine-tuning~\cite{zhou2025latent}, or result in hallucinated high-frequency details with stacked video diffusion models~\cite{chen2024mvsplat360}, failing to faithfully reconstruct the scene (see Figs.~\ref{fig:teaser} and~\ref{fig:qualitative_mvsplat360}).

We introduce \paper, a principled solution for high-fidelity novel view synthesis in latent space radiance fields. Our approach preserves high-frequency details, while operating in a frozen VAE setting. The core insight is to keep the 3D representation in the low-frequency domain and recover high-frequency details in 2D-space from reference input views.
More specifically, we first encode each input view into latent space using a pre-trained VAE and optimize a 3DGS model on these latents. We show that this process suffers from poor geometric and photometric reconstruction quality due to multi-view inconsistencies. To enhance rendered latents from novel views, we leverage latent diffusion with multi-view attention~\cite{singer2022makeavideo,ho2022imagenvideo,blattmann2023stable,guo2023animatediff,chen2023videocrafter}, which recovers fine details by conditioning on nearby views. 
Importantly, our VAE remains frozen, preserving the reconstruction quality and generalization capacity of pretrained autoencoders, trained on billions of images~\cite{rombach2022high}. We further demonstrate that our proposed framework can enhance the results from MVSplat360~\cite{chen2024mvsplat360}, a feed-forward latent 3DGS model capable of novel view synthesis with as few as 5 input views. Our method consistently improves fine details while mitigating hallucinations. 
Extensive experiments across multiple datasets show that our method outperforms previous latent-based radiance field approaches in terms of image quality and consistency with source views. 


To summarize, our contributions are as follows:

\begin{itemize}
\item \textbf{An in-depth analysis of 3D reconstruction in VAE latent space.} We demonstrate that the latent spaces of multiple popular VAEs lack multi-view consistency, limiting their applicability for latent-based 3D reconstruction.

\item \textbf{A principled framework for latent 3D reconstruction.} We show that high-frequency detail preservation in latent space highly benefits from 2D-space context, introducing multi-view attention as a key mechanism.

\item \textbf{State-of-the-art latent radiance field reconstruction.} \paper significantly outperforms existing latent-based methods~\cite{zhou2025latent}, achieving superior quality on dense, sparse and cross-dataset generalization tasks.

\item \textbf{Integration with feed-forward 3D models.} We show compatibility with existing latent-based feed-forward frameworks like MVSplat360~\cite{chen2024mvsplat360}, demonstrating consistent improvements in sparse-view scenarios with minimal modification.

\end{itemize}
\section{Related Work} \label{sec:related}

\begin{figure*}[t]
\centering
\includegraphics[trim={0 0 0 0}, clip, width=\textwidth]{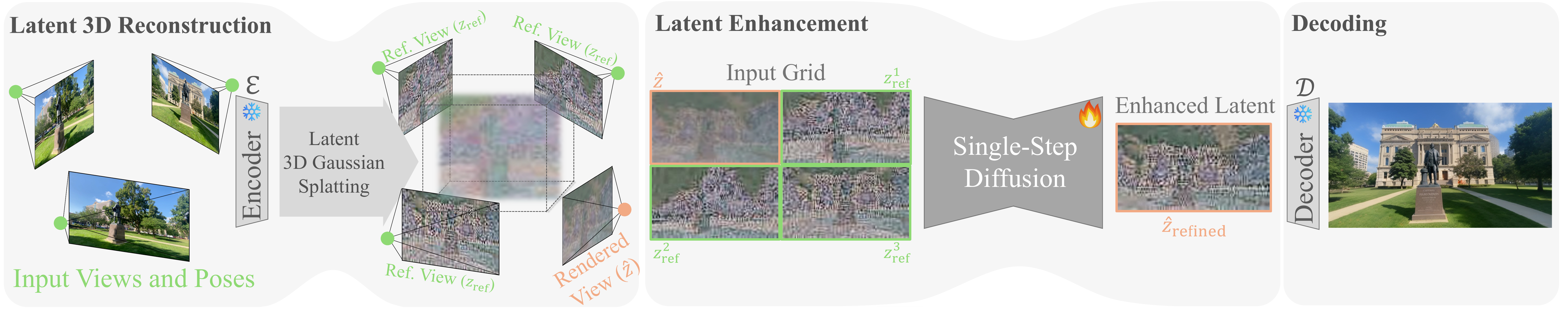} \\
\caption{\textbf{Framework Overview}. 
Given a set of input views with known camera parameters, each image is encoded into the VAE latent space of a diffusion model. We then perform 3DGS optimization to reconstruct the underlying latent radiance field.
Due to multi-view inconsistencies in diffusion VAEs latent space, a rendered novel view latent lacks high frequency details. We tile this rendered view together with reference views into a grid, and leverage a single-step diffusion model with self-attention mechanism that aggregates information across all views. The enhanced latent image is finally decoded to receive the novel view image.
}
\label{fig:architecture}
\end{figure*}

\subsection{Latent Radiance Fields}
\paragraph{Radiance fields.} Neural Radiance Fields (NeRFs)~\cite{mildenhall2020nerf} revolutionized 3D reconstruction by representing scenes as continuous volumetric functions. Subsequent work has explored various extensions including faster training~\cite{muller2022instant,chen2022tensorf}, unbounded scenes~\cite{barron2022mip}, and dynamic content~\cite{pumarola2021dnerf,park2021nerfies}. 3D Gaussian splatting~\cite{kerbl20233dgaussians} introduced an explicit point-based representation with differentiable rasterization, enabling real-time rendering. Recent methods have extended Gaussian splatting to dynamic scenes~\cite{luiten2023dynamic}, large-scale environments~\cite{xu2024streetgaussians}, and feed-forward prediction~\cite{chen2024mvsplat,charatan2024pixelsplat}. 

\paragraph{VAE latent space radiance fields.} 
Recent work has explored operating in compressed latent spaces, typically used by diffusion models, for 3D reconstruction and rendering. This latent space offers computational efficiency, and natively integrates into latent-diffusion pipelines.
For NeRFs, several methods~\cite{metzer2023latent, park2023ed} train directly in the latent space for 3D generation.
3DGS-based approaches have adopted a similar approach. Feature-3DGS~\cite{zhou2024feature} distills 3D feature fields from 2D foundation models using Gaussian splatting to render 3D-consistent features. By assigning learnable feature parameters to each Gaussian, features can be rendered for arbitrary views. This approach generalizes to any latent space, including the VAE latent space we focus on. 
Building on this, latentSplat~\cite{wewer2024latentsplat} extended 3DGS to address the uncertainty and generative nature of novel view synthesis given only two input views. They assigned each Gaussian with VAE mean and variance parameters, allowing for sampling during the splatting procedure. 
Later, MVSplat360~\cite{chen2024mvsplat360} suggested a feed-forward pipeline which directly predicts and renders VAE features, and refines them by using video diffusion for consistent novel view synthesis. 
Recently, Latent Radiance Fields (LRF)~\cite{zhou2025latent} was the first to observe that although useful, this latent space exhibits non-3D consistent characteristics. They fine-tuned the VAE, making it more 3D-consistent between different views. However, fine-tuning the VAE introduces degraded reconstructions and makes it harder to integrate into already pre-trained diffusion models, that expect the former input latent distribution.

\paragraph{Diffusion models for 3D generation.}
Diffusion models~\cite{ho2020denoising,song2021scorebased} have demonstrated remarkable generative capabilities~\cite{rombach2022high}. Recent advances have successfully integrated diffusion models with 3D representations, enabling applications such as text-to-3D generation~\cite{poole2022dreamfusion,lin2023magic3d,tang2023dreamgaussian}, novel view synthesis~\cite{watson2022novel,liu2023zero1to3}, and 3D reconstruction~\cite{kerr2023lerf,wu2023reconfusion}. Several methods leverage diffusion for multi-view consistent generation. Difix3D~\cite{wu2025difix3d} employs a fine-tuned single-step diffusion model to enhance rendered RGB views using reference non-distorted views. DiffusioNeRF~\cite{wynn2023diffusionerf} and ReconFusion~\cite{wu2023reconfusion} apply diffusion priors to ensure consistency across multiple viewpoints.

To aggregate multi-view information in 3D tasks, attention mechanisms have proven highly effective. Transformers have been successfully applied to multi-view stereo~\cite{ding2022transmvsnet}, novel view synthesis~\cite{johari2022geonerf}, and feed-forward 3D prediction~\cite{szymanowicz2023splatter,charatan2024pixelsplat}. Cross-attention, in particular, enables effective information flow between views~\cite{shi2023gnt,wang2023ibrnet}, and recent diffusion-based approaches leverage attention mechanisms to enforce multi-view consistency~\cite{shi2023mvdream,liu2023syncdreamer}. While existing methods typically perform diffusion in the VAE latent space but subsequently model 3D in RGB space, our approach conducts the entire pipeline—rendering and enhancement—entirely in latent space, utilizing multi-view attention during diffusion to fuse rendered features with nearby input views for detail recovery.

\section{Preliminary}
\label{sec:pre}

\begin{figure*}[ht]
\centering
\begin{minipage}{0.59\textwidth}
    \centering
    \includegraphics[width=\textwidth]{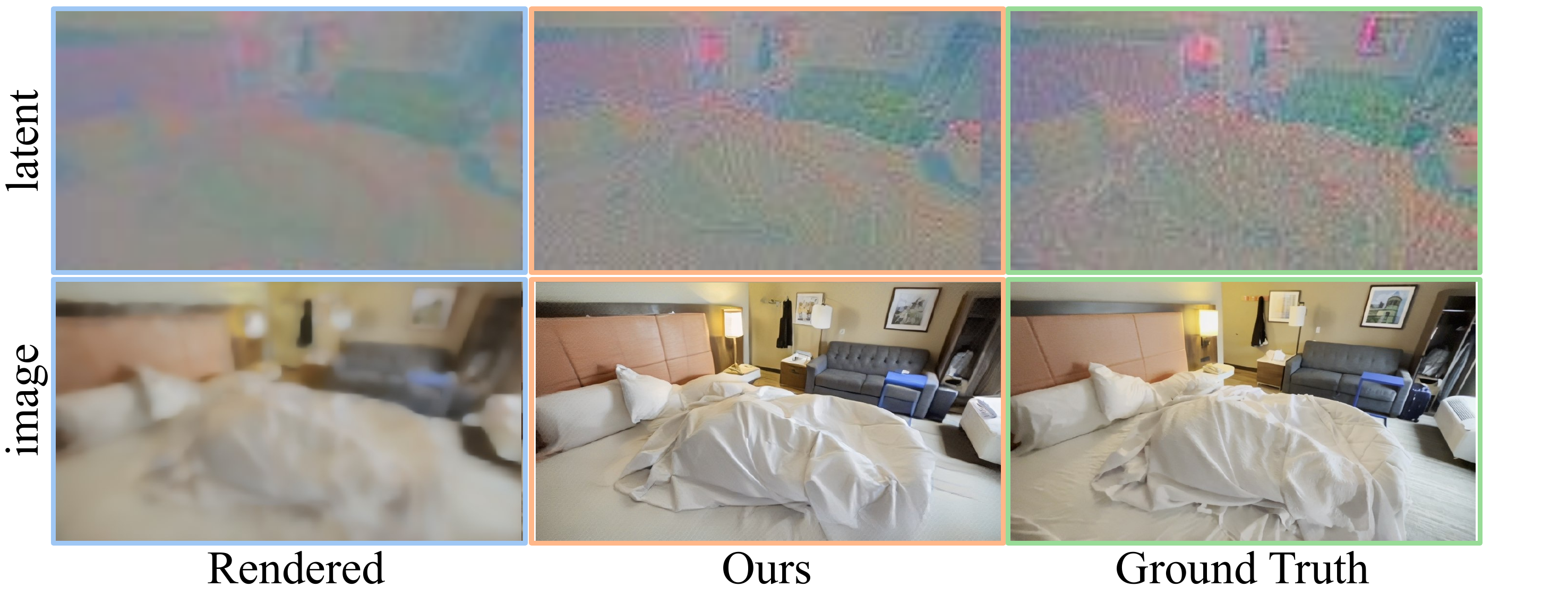}
    \caption*{(a)}
\end{minipage}\hfill
\begin{minipage}{0.4\textwidth}
    \centering
    \includegraphics[width=\textwidth]{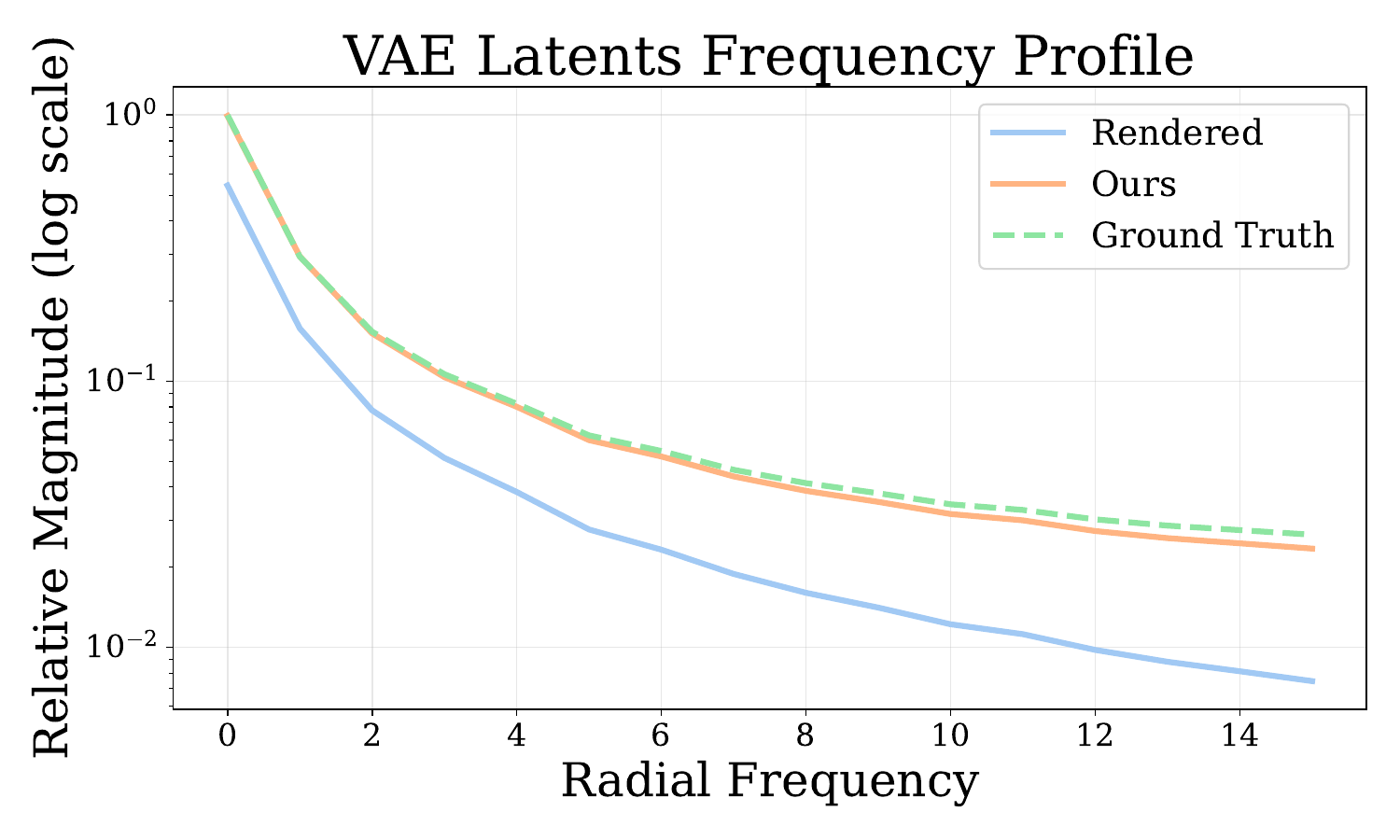}
    \caption*{(b) }
\end{minipage}
\caption{\textbf{VAE latents spectral analysis.} (a) Images in latent space and the corresponding image space (after decoding) (b) Magnitude spectrum of the latent image (Rendered, Ours and Ground Truth), normalized to $1$. In both visualizations, VAE latents contain both low- and high-frequency components (green). During 3DGS optimization, inconsistent high frequencies average out, leaving only low-frequency components (blue) and causing blurry decoded images. Our method produces latents whose spectrum closely matches that of the original VAE latents, reconstructing high-frequency details (orange). Graphs show averages over more than 45K latent images from 140 scenes.}
\label{fig:spectral_all_channels}
\end{figure*}

\paragraph{3D Gaussian splatting.} 


3D Gaussian splatting~\cite{kerbl20233dgaussians} represents a scene using a set of 3D Gaussians, which can be rendered into an image from a given camera $C$.
Each Gaussian is parameterized as
\begin{equation}
\label{eq:3dgs}
G = (\mu, \Sigma, \alpha, f_{\text{c}}),
\end{equation}
where the spatial properties are defined by the mean $\mu \in \mathbb{R}^3$ (center position) and covariance $\Sigma \in \mathbb{R}^{3 \times 3}$ (shape and orientation). Together, these determine the 3D extent and anisotropic structure of each Gaussian. The scalar opacity $\alpha \in \mathbb{R}$ controls blending during rendering, allowing soft compositing of overlapping Gaussians. 
Each Gaussian is also characterized with a color representation, $f_c$, which can be RGB color values or spherical harmonics~\cite{kerbl20233dgaussians, fridovich2022plenoxels}.
During rendering, these colors are alpha-composited via differentiable splatting to produce an output image.

\paragraph{VAEs in latent diffusion models.}
The typical approach for diffusion-based generation involves initially transforming RGB images into a compressed latent representation~\cite{rombach2022high}. This transformation is achieved through an autoencoder that maps the pixel space to either continuous or discrete latent codes. We focus on the continuous variant, which is more widely adopted in practice. For a given image $I \in \mathbb{R}^{H \times W \times 3}$, an encoder network $\mathcal{E}$ maps the image to a lower-dimensional latent code
\begin{equation}
\label{eq:latent}
z = \mathcal{E}(I) \in \mathbb{R}^{h \times w \times d},
\end{equation}
where $h = \frac{H}{f}$, $w = \frac{W}{f}$, $f$ indicates the downsampling factor and $d$ is the latent dimension. A decoder network $\mathcal{D}$ then takes this latent and recovers the original image as

\begin{equation}
\label{eq:decode}
\hat{I} = \mathcal{D}(z). 
\end{equation}

\section{Method}

\label{sec:method}
Our approach for novel view synthesis consists of two main stages. In the first stage, we extract a 3D Gaussian splatting representation given some input images, directly in the VAE latent space. However, as we will show, this step alone is insufficient for achieving high-fidelity reconstructions due to inconsistencies across multi-view latents. To address this limitation, the second stage introduces a diffusion-based refinement mechanism that leverages attention to fuse rendered latent features from 3DGS with source input view latents during the denoising process. This diffusion-aware fusion leads to more consistent and higher-quality results.

In the following sections, we detail the 3DGS formulation operating in the latent space (Section~\ref{sec:3DGS_latents}), analyze the reconstruction challenges inherent to this setup (Section~\ref{sec:vae_mv_inconsistencies}) and describe our diffusion-based refinement strategy for rendered latents (Section~\ref{sec:diffusion_aware_reference}). The overall pipeline is illustrated in Fig.~\ref{fig:architecture}.

\subsection{Latent 3D Gaussian Splatting}
\label{sec:3DGS_latents}
Similar to Feature-3DGS~\cite{zhou2024feature}, we adapt 3DGS to operate in a feature space rather than directly in image color space. We use the VAE latent representation as our feature space, as also done by~\cite{zhou2025latent}. Specifically, the 3D Gaussians parameters in Eq.~\ref{eq:3dgs} are extended with additional $f_z\in\mathbb{R}^d$ values. By splatting the Gaussians, we can render views in the latent space.

Given a set of input views $\{I_i\}_{i=1}^N$ with known camera parameters $\{C_i\}_{i=1}^N$, each image is encoded into latent space via Eq.~\ref{eq:latent}, producing $\{z_i\}_{i=1}^N$. We then perform 3DGS optimization to reconstruct the underlying latent radiance field. A rendered novel view latent $\hat{z}$ can be decoded back into image space using Eq.~\ref{eq:decode}.
This same reconstruction can also be obtained through a feed-forward approach~\cite{chen2024mvsplat360}. In this case, 
a feed-forward network directly predicts the latent 3DGS representation from input views. 


\subsection{Multi-View Inconsistencies in VAE Latents}
\label{sec:vae_mv_inconsistencies}
Our work is driven by the observation that existing VAE models, such as Stable Diffusion VAE~\cite{rombach2022high}, produce latent representations that lack 3D consistency, limiting their direct use for 3D scene reconstruction and novel view synthesis. 
This limitation arises from two related spectral deficiencies. First, the latent spaces fail to maintain equivariance under basic spatial transformations like scaling and rotation~\cite{kouzelis2025eq}. Second, and more importantly, view-dependent high-frequency components, essential for accurate decoding, exhibit the most severe 3D inconsistencies across viewpoints, unlike in RGB space~\cite{skorokhodov2025improving}.


When optimizing 3D Gaussian splatting in latent space, this spectral inconsistency becomes particularly problematic. As also observed by LRF~\cite{zhou2025latent}, high-frequency components are highly view-dependent and fail to agree across training views. This leads to conflicting signals during the optimization process, effectively cancelling out the high frequencies. Consequently, the latent space retains only coarse structure, while losing the fine details required for photorealistic rendering, leading to blurred outputs.
This effect is demonstrated in Fig.~\ref{fig:spectral_all_channels}, where rendered latent features exhibit significantly attenuated high-frequency content compared to the original encoded features, as seen in the reference and ground-truth views.
We further show in the Appendix that this phenomenon is widespread across other VAE models. In contrast, our approach produces latents that better preserve high-frequency components, as shown in Fig.~\ref{fig:spectral_all_channels}, resulting in spectra that more closely match the original VAE latent representation.

\subsection{Diffusion-Based Latent Refinement}
\label{sec:diffusion_aware_reference}
To address the lack of high-frequency details in rendered latent features, we propose a diffusion-based refinement module that enhances the rendered latent. 
Inspired by~\cite{wu2025difix3d}, our method leverages reference views to recover missing details while preserving the geometry of the rendered latent. Unlike their work, which focuses on correcting artifacts in image-space renderings, we aim to reconstruct lost details arising from 3D inconsistencies in VAE latents.

We base our model on a single-step diffusion model, which achieves efficient performance during inference. To enable effective cross-view information transfer, we condition the diffusion model on reference views by arranging the inputs in a spatial grid, following~\cite{
winter2025objectmate,kang2025latentunfold,Li2025,Zheng2024Analogist,Oorloff2025}. This grid-based arrangement has been shown to be effective at object preservation across views. This approach offers a clean solution for information sharing without requiring diffusion model-specific architectural changes, enabling our method to work with future diffusion models.
The input grid contains latents extracted from reference images, with the degraded latent placed in the top-left corner (see Fig.~\ref{fig:architecture}). Reference views are selected as the closest training views to the degraded latent in both position and orientation. During denoising, the attention mechanism propagates high-frequency details from the references to the rendered latent, mitigating artifacts and improving reconstruction quality. Alternative strategies for injecting reference views are presented in the Appendix.


Formally, for a rendered latent $\hat{z}$, we combine $V$ additional reference latents $\{z^i_{\text{ref}}\}_{i=1}^V$, encoded from nearby training views. All latents are tiled into a grid $\hat{z}_\text{grid} \in \mathbb{R}^{(V+1) \times M \times d}$, where $d$ is the number of latent channels and $M = h \times w$. The view axis is then merged into the spatial dimension, resulting in $z \in \mathbb{R}^{((V+1) \cdot M) \times d}$, and self-attention is applied jointly across all views. The resulting grid is passed to the diffusion model, which outputs a refined latent grid, from which we take the top-left position as the enhanced latent, $\hat{z}_\text{refined}$. This enhanced latent is then decoded back to image space with Eq.~\ref{eq:decode}

\paragraph{Training objective.} 
During training, we use rendered latents $\hat{z}$ from known training cameras, for which we also have the corresponding encoded ground-truth latents $z_\text{gt}$. The refinement is supervised by comparing the enhanced latent with the ground-truth
\begin{equation}
    \mathcal{L}_{\text{recon}} = \| \hat{z}_\text{refined} - z_\text{gt} \|_2^2.
\end{equation}
To improve perceptual quality, we include LPIPS~\cite{zhang2018perceptual} and RGB reconstruction losses on decoded images

\begin{align}
    \mathcal{L}_{\text{LPIPS}} &= \text{LPIPS}\big(\mathcal{D}(\hat{z}_\text{refined}), \mathcal{D}(z_\text{gt})\big), \\
    \mathcal{L}_{\text{RGB}} &= \big\| \mathcal{D}(\hat{z}_\text{refined}) - \mathcal{D}(z_\text{gt}) \big\|_2^2,
\end{align}
where $\mathcal{D}$ denotes the VAE decoder. The total training loss is
\begin{equation}
    \label{eq:loss}
    \mathcal{L}_{\text{total}} = \mathcal{L}_{\text{recon}} + \lambda_{\text{LPIPS}} \mathcal{L}_{\text{LPIPS}} + \lambda_{\text{RGB}} \mathcal{L}_{\text{RGB}}.
\end{equation}
This formulation emphasizes that the model enhances a degraded latent frame by leveraging both the 3DGS rendering and nearby reference views through self-attention in the diffusion model.

\begin{figure*}[t]
\centering
\setlength{\tabcolsep}{2pt} 
\begin{tabular}{@{}cccc@{}}
\makebox[0.21\textwidth][c]{\small\textbf{Feature-3DGS}} & 
\makebox[0.21\textwidth][c]{\small\textbf{LRF}} & 
\makebox[0.21\textwidth][c]{\small\textbf{\paper (Ours)}} & 
\makebox[0.21\textwidth][c]{\small\textbf{Ground Truth}} \\[2pt]

\includegraphics[width=0.21\textwidth]{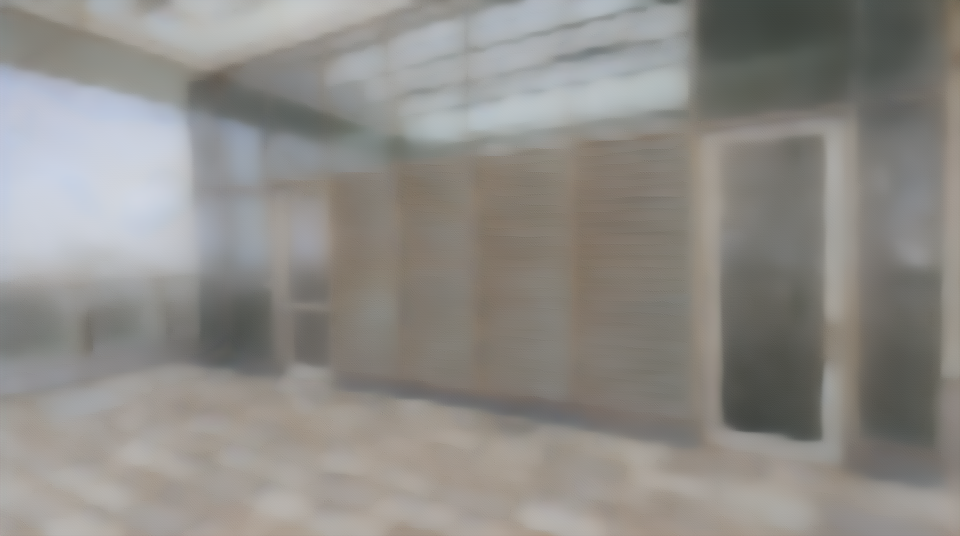} &
\includegraphics[width=0.21\textwidth]{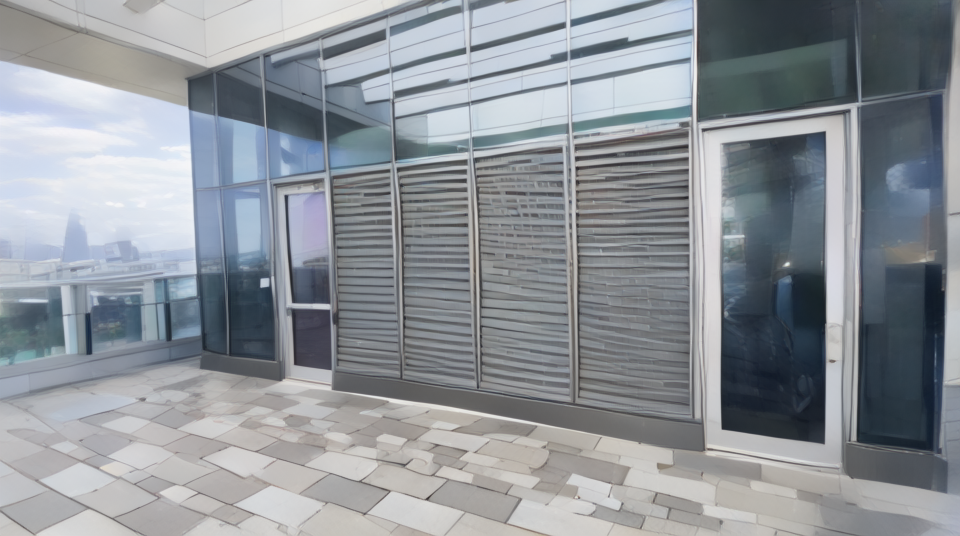} &
\includegraphics[width=0.21\textwidth]{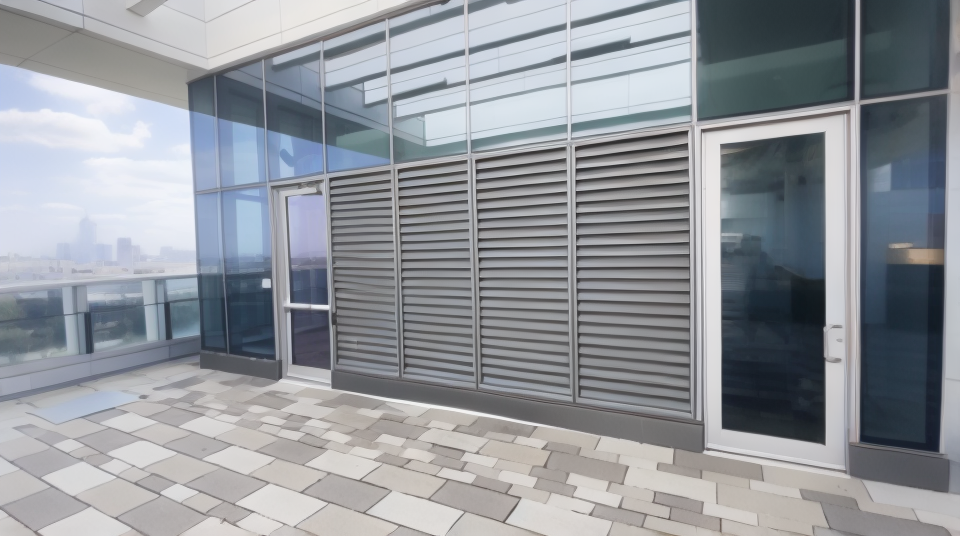} &
\includegraphics[width=0.21\textwidth]{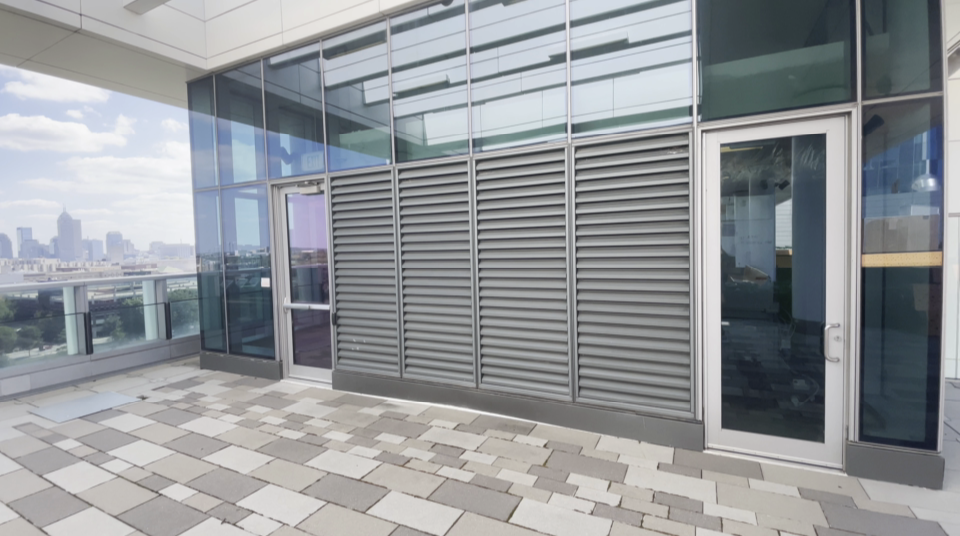} \\

\includegraphics[width=0.21\textwidth]{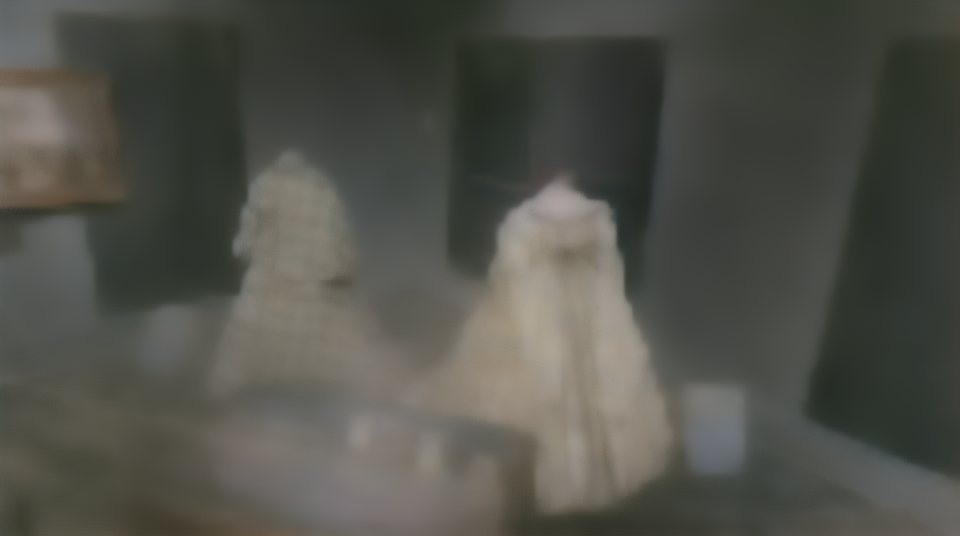} &
\includegraphics[width=0.21\textwidth]{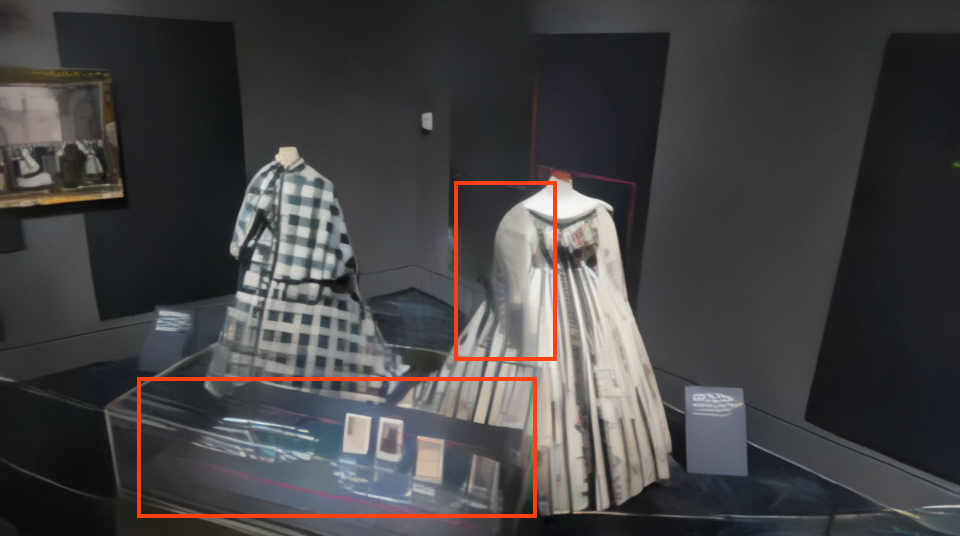} &
\includegraphics[width=0.21\textwidth]{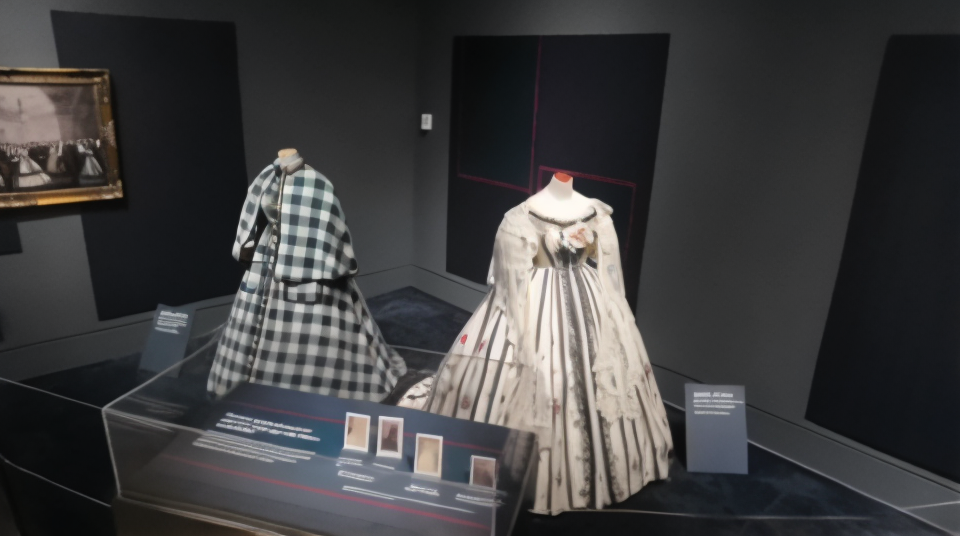} & 
\includegraphics[width=0.21\textwidth]{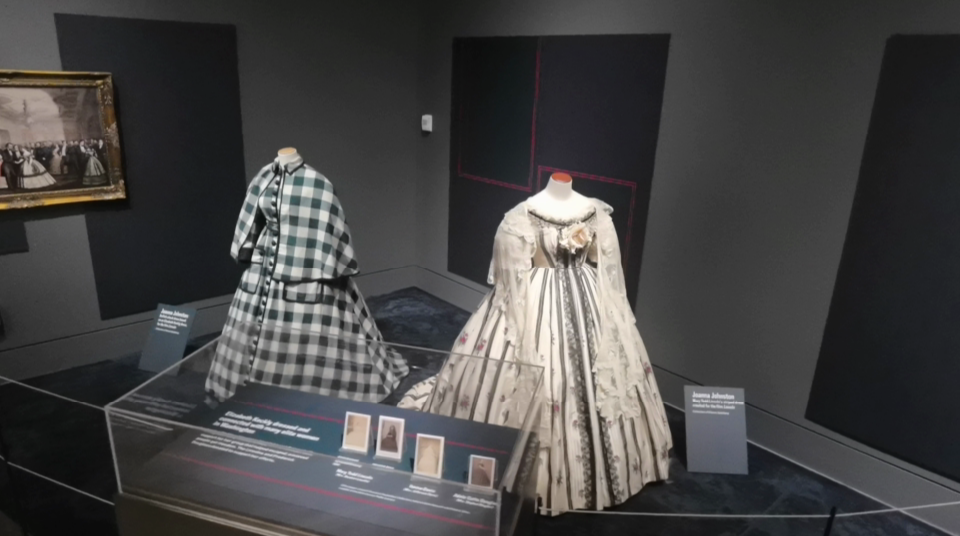} \\

\includegraphics[width=0.21\textwidth]{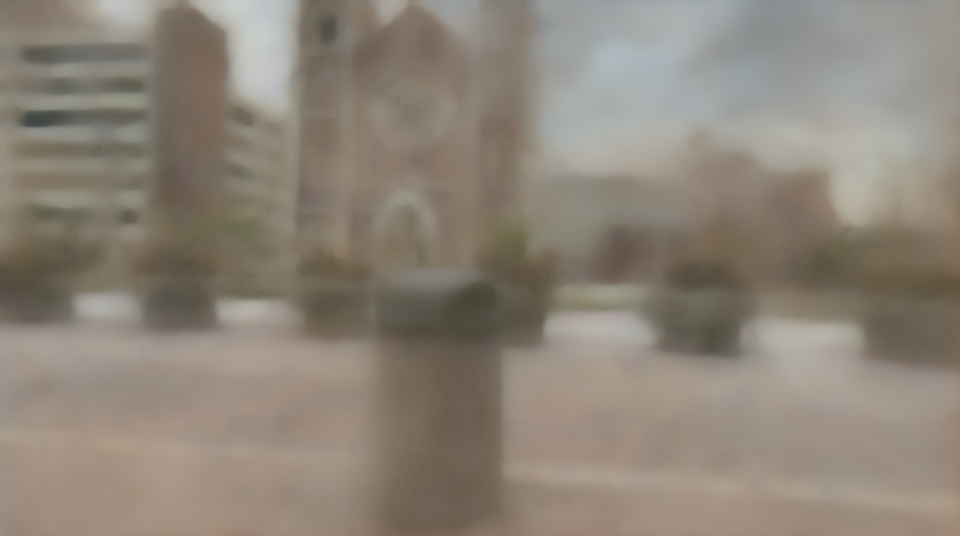} &
\includegraphics[width=0.21\textwidth]{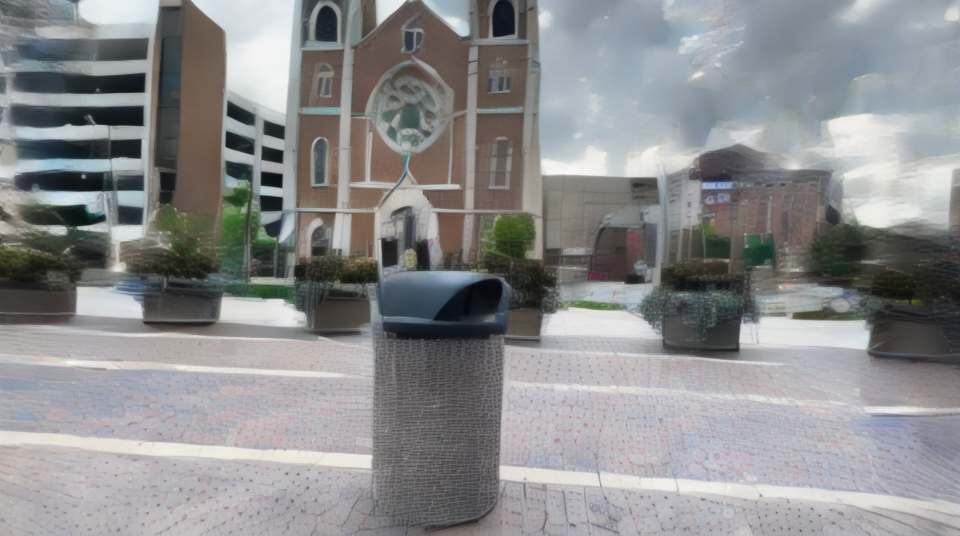} &
\includegraphics[width=0.21\textwidth]{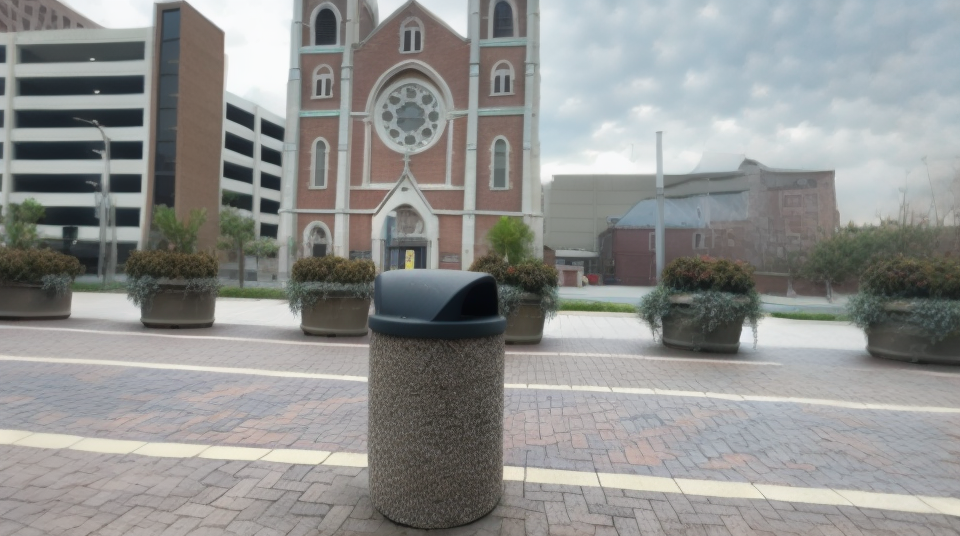} &
\includegraphics[width=0.21\textwidth]{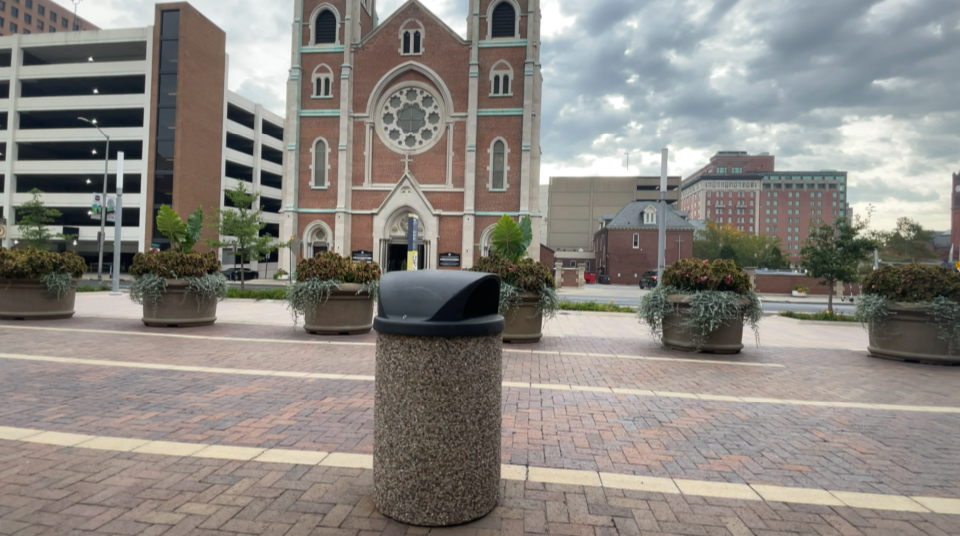}
\\

\includegraphics[width=0.21\textwidth]{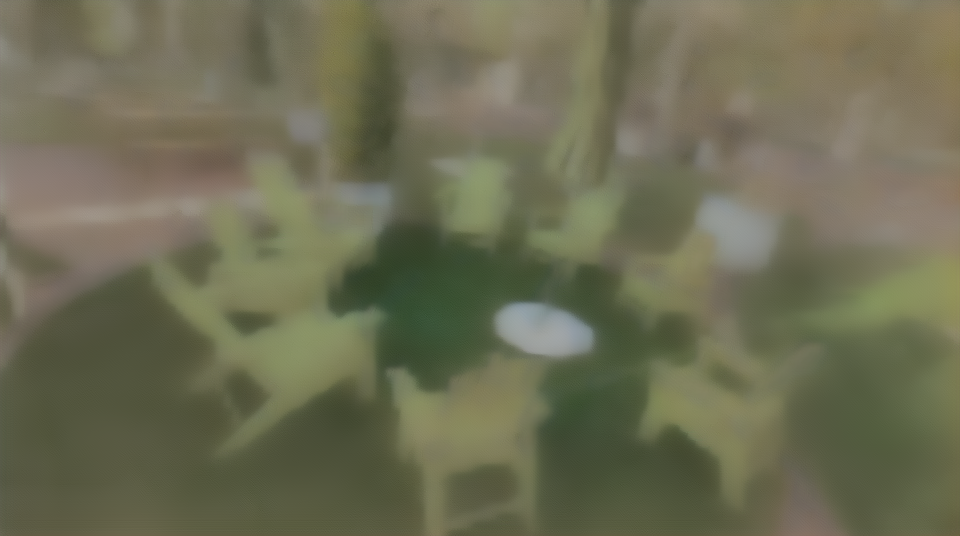} &
\includegraphics[width=0.21\textwidth]{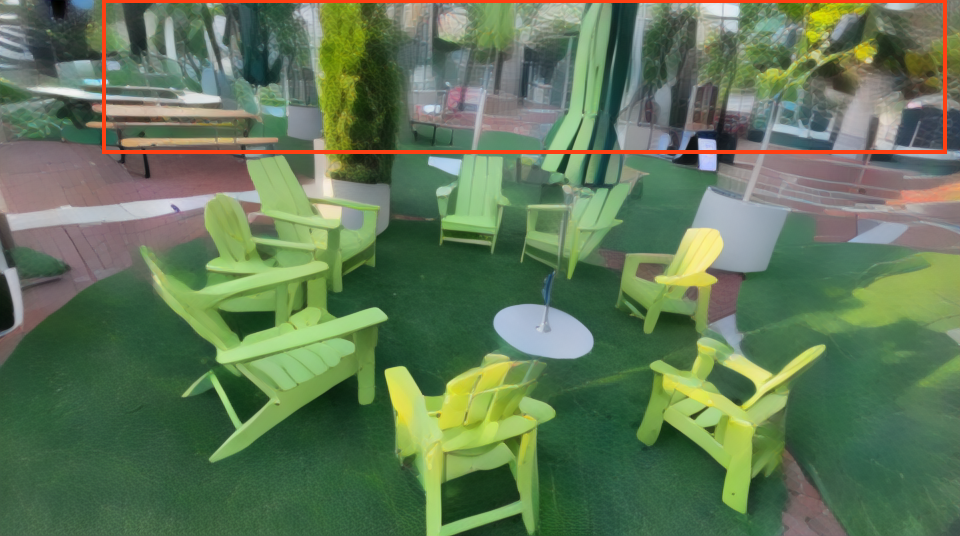} &
\includegraphics[width=0.21\textwidth]{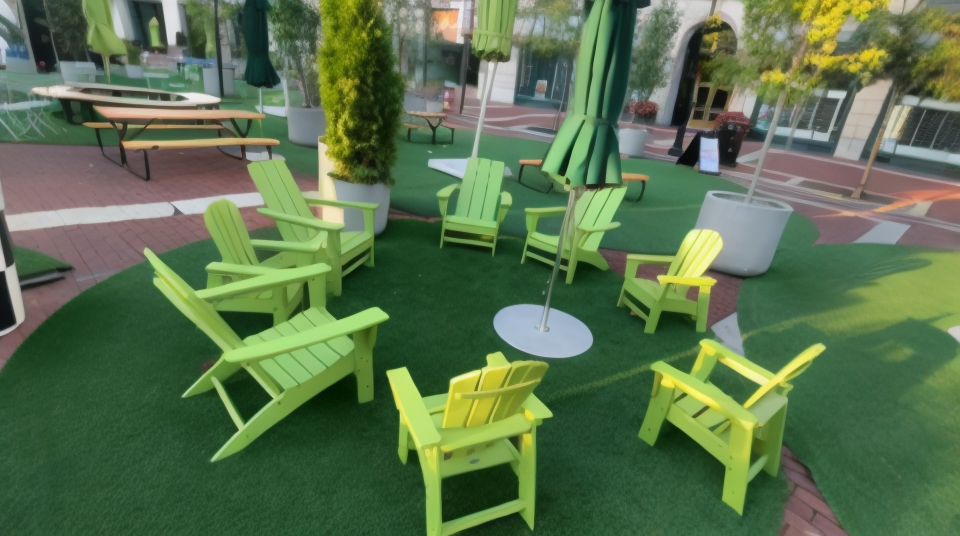} &
\includegraphics[width=0.21\textwidth]{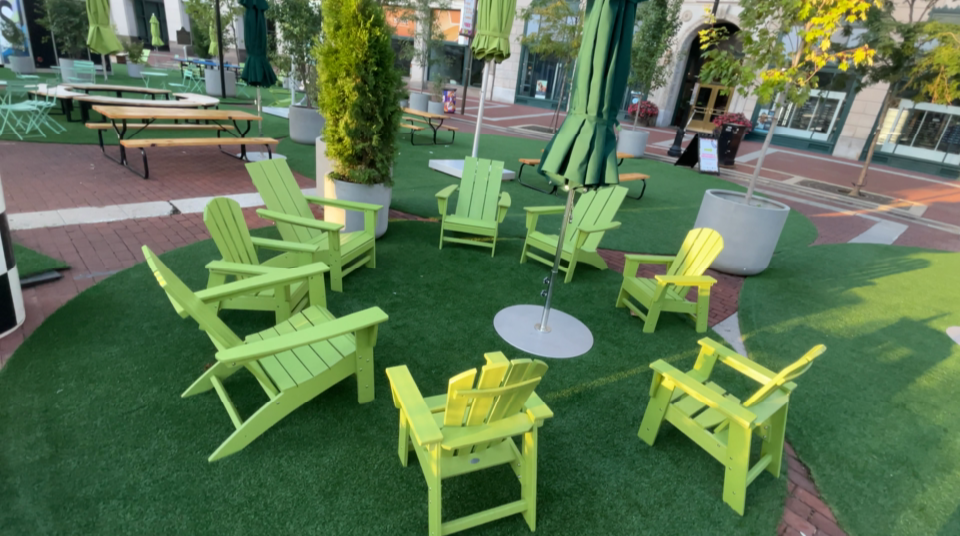} \\
\end{tabular}
\caption{\textbf{Qualitative comparison.} We compare \paper to other latent radiance field methods on novel view synthesis reconstruction quality. Feature-3DGS~\cite{zhou2024feature} exhibits considerable loss of detail, and LRF~\cite{zhou2025latent} improves upon this baseline but still fails to recover fine details. In contrast, \paper produces sharper and more faithful reconstructions. The scenes are taken from the DL3DV-10K dataset.}

\label{fig:qualitative}
\end{figure*}

\section{Experiments}
\label{sec:experiments}

Our method consists of a latent space 3DGS reconstruction stage followed by diffusion-based refinement. Novel view latents are extracted and fed into our diffusion model for refinement. We evaluate quantitatively and qualitatively the reconstructed novel views. In addition, we demonstrate the effectiveness of \paper as an enhancer in a feed-forward latent 3DGS setting.


\begin{table*}[t]
\centering
\caption{\textbf{Quantitative comparison.} We compare our method using DL3DV-10K, LLFF and Mip-NeRF360 datasets. In the dense setting, we use 30 input views (except for the LLFF dataset, for which we use $1/8$ of views in each scene). In the sparse setting, we use 5 input views. The rest of the views in each scene are used for evaluation. LRF and \paper are trained only on DL3DV-10K. Best results in \textbf{bold}.}
\label{tab:main_results}
\resizebox{\textwidth}{!}{
\begin{tabular}{c|l|cccc|cccc|cccc}
\toprule
& & \multicolumn{4}{c}{\textbf{DL3DV-10K}} & \multicolumn{4}{c}{\textbf{LLFF}} & \multicolumn{4}{c}{\textbf{Mip-NeRF360}} \\
\cmidrule(lr){3-6} \cmidrule(lr){7-10} \cmidrule(lr){11-14}
& Method & PSNR$\uparrow$ & SSIM$\uparrow$ & LPIPS$\downarrow$ & FID$\downarrow$ & PSNR$\uparrow$ & SSIM$\uparrow$ & LPIPS$\downarrow$ & FID$\downarrow$ & PSNR$\uparrow$ & SSIM$\uparrow$ & LPIPS$\downarrow$ & FID$\downarrow$ \\
\midrule
\multirow{3}{*}{\rotatebox[origin=c]{90}{Dense}} & Feature-3DGS & 16.37 & 0.545 & 0.704 & 263.45 &
16.23 & 0.520 & 0.644 & 257.20 & 
14.85 & 0.417 & 0.739 & 294.82 \\
& LRF~\cite{zhou2025latent} & 20.19 & 0.619 & 0.322 & 75.32 & 
17.98 & 0.542 & 0.379  & 103.18 & 
19.08 & 0.489 & 0.409 & 135.22 \\
& \textbf{\paper (Ours)} & \textbf{21.94} & \textbf{0.692} & \textbf{0.265} & \textbf{35.60} & 
\textbf{19.57} & \textbf{0.610} & \textbf{0.307 } & \textbf{52.14} & 
\textbf{20.42} & \textbf{0.546} & \textbf{0.364} & \textbf{70.90} \\

\midrule
\multirow{3}{*}{\rotatebox[origin=c]{90}{Sparse}} & Feature-3DGS & 15.04 & 0.519 & 0.742 & 308.00 &
15.71 & 0.512 & 0.660 & 273.26 & 
14.35 & 0.406 & 0.738 & 314.36 \\
& LRF~\cite{zhou2025latent} & 15.34 & 0.488 & 0.494 & 204.36 &
17.86 & 0.529 & 0.387  & 102.57 & 
14.25 & 0.329 & 0.603 & 310.76 \\
& \textbf{\paper (Ours)} & \textbf{17.44} & \textbf{0.573} & \textbf{0.429} & \textbf{86.12} &
\textbf{18.53} & \textbf{0.592} & \textbf{0.362} & \textbf{76.62} & 
\textbf{16.701} & \textbf{0.450} & \textbf{0.501} & \textbf{127.93} \\

\bottomrule
\end{tabular}
}
\end{table*}

\begin{figure*}[t]
\centering
\begin{tabular}{@{}ccc@{}}
\makebox[0.25\textwidth][c]{\small\textbf{MVSplat360}} & 
\makebox[0.25\textwidth][c]{\small\textbf{MVSplat360 + \paper (Ours)}} & 
\makebox[0.25\textwidth][c]{\small\textbf{Ground Truth}} \\[2pt]

\includegraphics[width=0.25\textwidth]{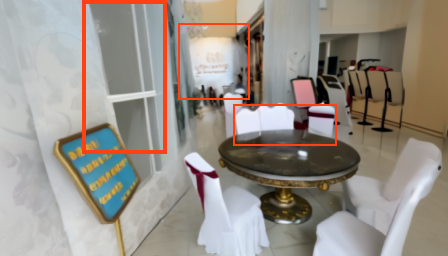} &
\includegraphics[width=0.25\textwidth]{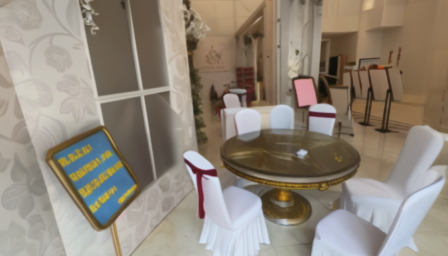} &
\includegraphics[width=0.25\textwidth]{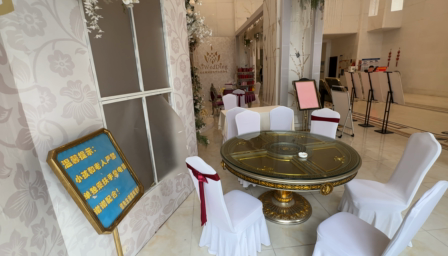} \\

\includegraphics[width=0.25\textwidth]{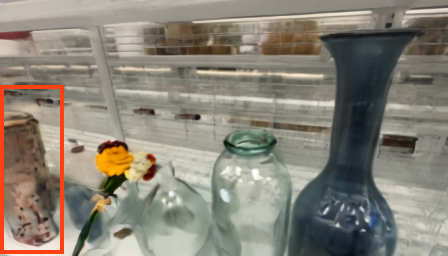} &
\includegraphics[width=0.25\textwidth]{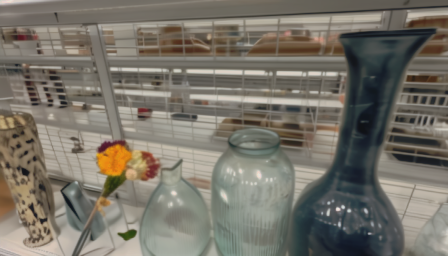} &
\includegraphics[width=0.25\textwidth]{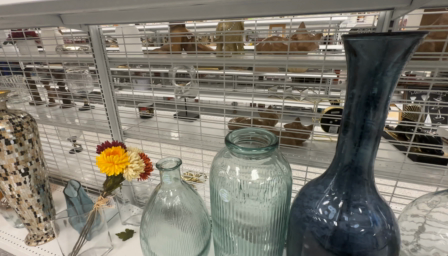} \\

\includegraphics[width=0.25\textwidth]{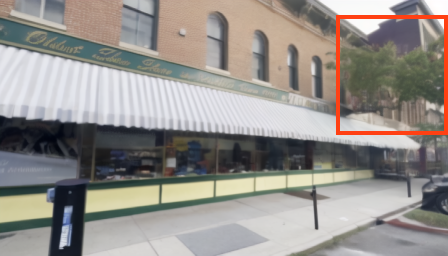} &
\includegraphics[width=0.25\textwidth]{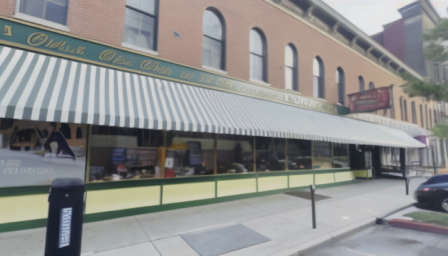} &
\includegraphics[width=0.25\textwidth]{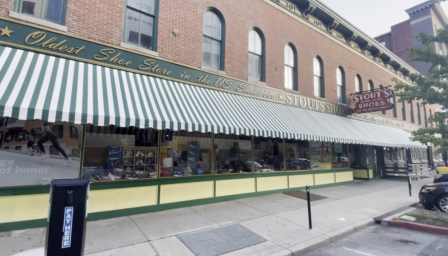} \\

\end{tabular}
\vspace{-1mm}
\caption{\textbf{Feed-Forward Qualitative comparison}. We demonstrate how \paper can enhance feed-forward latent radiance field methods such as MVSplat360~\cite{chen2024mvsplat360}. While MVSplat360 often hallucinates (e.g., the window in the first example or the tree in the last example) and lacks fine details, \paper yields sharper and more faithful reconstructions.}
\label{fig:qualitative_mvsplat360}
\end{figure*}


\paragraph{Baselines.} We compare to Feature-3DGS~\cite{zhou2024feature} using the original implementation but without the encoder-decoder channel compression modules. Since the VAE latent has only 4 channels, channel compression is unnecessary. 
We also compare to LRF~\cite{zhou2025latent}, a method for 3D latent reconstruction. We use the original implementation to train and evaluate LRF with their default hyperparameters.

\paragraph{Metrics.} 
We evaluate our approach using standard metrics that assess reconstruction quality.
Specifically, we employ Peak Signal-to-Noise Ratio (PSNR) and Structural Similarity Index (SSIM)~\cite{wang2004image} to measure pixel-level accuracy and structural fidelity of the reconstructed scenes. Additionally, we use Learned Perceptual Image Patch Similarity (LPIPS)~\cite{zhang2018perceptual} and Fréchet Inception Distance (FID)~\cite{heusel2017gans} to evaluate perceptual quality and distribution similarity between rendered and ground truth images.


\subsection{Datasets}
\paragraph{DL3DV-10K~\cite{ling2024dl3dv}.} A challenging dataset of real-world scenes. 
It includes 10K scenes providing diverse indoor and outdoor environments with varying lighting conditions and complex geometries. The benchmark consists of 140 scenes. Importantly, the training and benchmark splits overlap, so we filter the benchmark scenes from our training set.
We use a $960 \times 540$ resolution for efficient optimization, though our method is not limited to this resolution.

\paragraph{Generalization datasets.} 
To demonstrate generalization capabilities, we evaluate on two additional standard benchmarks. The Mip-NeRF360 dataset~\cite{barron2022mip} contains 9 scenes (5 outdoor and 4 indoor) featuring complex central objects with detailed backgrounds, captured under controlled conditions with fixed camera exposure and minimal lighting variation. The LLFF dataset~\cite{mildenhall2019local} consists of 8 forward-facing scenes, containing 20-62 images per scene. 



\subsection{Implementation details}




\paragraph{3D Gaussian splatting.} We optimize latent 3DGS representations for each scene with dense and sparse input view configurations: 30 views in the dense setting (for LLFF, we sample $1/8$ of available views), and 5 in the sparse setting. To ensure optimal spatial coverage, we sample camera positions using furthest point sampling.

\paragraph{Diffusion model.} We use the pre-trained KL-based VAE (with compression ratio $f = 8$) from Latent Diffusion Model~\cite{rombach2022high} and employ a pre-trained Stable Diffusion Turbo~\cite{sauer2024adversarial} for latent refinement, which enables fast, single-pass inference.
We set $V = 3$ reference views in the input grid to our diffusion model, selected from the set used for 3DGS optimization. The remaining novel views in each scene are used for supervision and evaluation. 
We fine-tune the diffusion model on a randomly selected subset of 400 scenes from the DL3DV-10K train split using 8 NVIDIA H100 GPUs for approximately 24 hours. We use the AdamW optimizer at a base learning rate of $2 \cdot 10^{-5}$, noise level $\tau=300$, and loss weights $\lambda_{\text{LPIPS}}=2$ and $\lambda_{\text{RGB}}=1$. The Appendix includes a robustness analysis of our model to the noise level $\tau$, results with varying hyperparameters, and runtime and memory analysis.

\subsection{Results}

\paragraph{Qualitative.} We show a qualitative comparison between Feature-3DGS, LRF and our approach in Fig.~\ref{fig:qualitative}. As shown, decoded novel views from Feature-3DGS lack fine-grained details and textures due to the multi-view inconsistencies of the latent space. In comparison, LRF reconstructs the scene more reliably, but trades latent space 3D consistency for reconstruction quality, resulting in loss of details. We inject details from reference views, leading to highly detailed and structurally grounded rendered images. Additional visual results can be found in the Appendix.


\paragraph{Quantitative.} Table~\ref{tab:main_results} presents our results across various settings and datasets, including experiments with extremely sparse input views (5 views) to demonstrate the robustness of our approach. Our method consistently outperforms existing approaches across all metrics, with particularly strong generalization on LLFF and Mip-NeRF360. 
This superior performance, especially under sparsity, stems from leveraging three complementary sources of information: prior knowledge from the diffusion model, fine-grained details from reference views and rough geometric structure from the rendered latents. 
This combination enables robust reconstruction where traditional methods fail due to insufficient geometric constraints, as the diffusion prior compensates for missing information while the reference views ensure accurate details grounded in the observed geometry.


We also asses the 3D-consistency of our method using MEt3R~\cite{asim24met3r}, a metric for measuring multi-view consistency.
\begin{table}
\centering
\caption{\textbf{3D Consistency.} MEt3R~\cite{asim24met3r} results (lower is better) on the DL3DV-10K dataset. Our method significantly outperforms both baselines.}
\begin{tabular}{lccc}
\hline
MEt3R$\downarrow$ & Feature-3DGS & LRF & \textbf{Splatent (Ours)} \\
\hline
Dense & 0.1106 & 0.1082 & \textbf{0.0774} \\
Sparse & 0.1281 & 0.1272 & \textbf{0.0998} \\
\hline
\end{tabular}
\label{tab:3d_consistency}
\end{table}
Results are shown in Table~\ref{tab:3d_consistency}. Splatent significantly outperforms both baselines, 28-30\% dense and  22\% sparse. We attribute this to our architectural design: the rendered latent provides geometric grounding (low-frequency structure), which is inherently consistent as it comes from 3D reconstruction. The diffusion model only needs to recover high-frequency details, which leads to more consistent detail completion across views.
We provide additional comparisons to diffusion-based methods in the Appendix.

\begin{table}[t]
\centering
\caption{\textbf{Components ablation.}
Impact of reference image count. Multiple references reduce hallucinations and enhance details, with performance saturating at 3 views.
}
\label{tab:ablation_attention}
\resizebox{0.48\textwidth}{!}{
\begin{tabular}{lcccc}
\toprule
Configuration & PSNR $\uparrow$ & SSIM $\uparrow$ & LPIPS $\downarrow$  & FID $\downarrow$\\
\midrule
No reference image & 19.47 & 0.626 & 0.389 & 83.66 \\
1 reference image & 21.61 & 0.683 & 0.276 & 38.04 \\
5 reference images & 21.96 & 0.692 & 0.263 & 35.16 \\
\midrule
\textbf{\paper (3 views)} & 21.94 & 0.692 & 0.265 & 35.60 \\
\bottomrule
\end{tabular}
}
\end{table}
\subsection{Ablation Studies}
\label{sec:ablation}

We perform an ablation study on the DL3DV-10K benchmark using the dense setting to evaluate key components in our method.
Table~\ref{tab:ablation_attention} reports the impact of using multiple reference images compared to a single view. Leveraging multiple references improves reconstruction quality, reducing hallucinations with better grounding of fine-grained scene details. Increasing the number of reference views further enhances performance but eventually saturates, as additional views contribute limited new information in the dense setting. We also note that memory consumption grows with the number of views. Thus, our base model uses 3 reference views, striking a balance between quality and efficiency. More ablations and visualizations are in the Appendix.

\subsection{Feed-Forward Setting}
We further demonstrate the effectiveness of our diffusion enhancement model for improving rendered image quality in a feed-forward latent-based 3DGS model. Specifically, we integrate \paper into MVSplat360~\cite{chen2024mvsplat360}, a feed-forward latent 3DGS method for sparse-view (5 input views) $360^{\circ}$ novel view synthesis. MVSplat360 renders Gaussian features directly into the latent space of a pre-trained Stable Video Diffusion (SVD) model, where these features guide the denoising process for video generation. 

We integrate our approach by applying the single-step diffusion  on the rendered latents, conditioned on reference latents extracted from the input views. This refinement occurs prior to the final SVD generation step, recovering attenuated high-frequency details before video synthesis. We train the entire model in an end-to-end fashion, where both our model and the SVD are fine-tuned. The weights of MVSplat360 are initialized with pre-trained weights, and the combined network is trained for additional 25K steps.

\paragraph{Results.} We demonstrate the effectiveness of \paper in improving MVSplat360 results in Fig.~\ref{fig:qualitative_mvsplat360}. While MVSplat360 leverages the generative prior of SVD to synthesize plausible content, fine-grained details remain blurred and are sometimes hallucinated. 
For example, the window in the top row or the tree in the bottom row in Fig.~\ref{fig:qualitative_mvsplat360} are not faithful to the structure of the scene, despite being covered by the input views.
Our integration accurately and faithfully recovers these details from reference views. This demonstrates that \paper addresses a complementary challenge: MVSplat360 benefits from strong diffusion priors for content generation, while \paper enforces fidelity and detail preservation relative to the input images.

\begin{table}
\centering
\caption{\textbf{Feed-forward latent 3DGS.} Quantitative results on DL3DV-10K using 5 input views. Our method consistently improves perceptual quality while maintaining geometric accuracy.}
\label{tab:mvsplat360}
\resizebox{0.48\textwidth}{!}{
\begin{tabular}{l|cccc}
\toprule
Method & PSNR$\uparrow$ & SSIM$\uparrow$ & LPIPS$\downarrow$ & FID$\downarrow$ \\
\midrule
MVSplat360~\cite{chen2024mvsplat360} & 16.691 & 0.514 & 0.431 & 13.462 \\
MVSplat360 + Splatent & \textbf{17.976} & \textbf{0.531} & \textbf{0.378} & \textbf{11.097} \\
\bottomrule
\end{tabular}
}
\end{table}

Moreover, Table~\ref{tab:mvsplat360} presents quantitative results showing consistent improvements across all metrics. By integrating our method, we achieve more perceptually faithful reconstructions with enhanced fine details and improved pixel-level accuracy.
These improvements pave the way for high-quality latent-based feed-forward methods and diffusion-based 3D reconstruction frameworks.

\section{Limitations}
While our method shows significant improvements in latent space radiance fields representations, some limitations remain. Our approach inherently faces a more challenging problem than RGB-space Gaussian splatting, as we operate in a lossy latent space where information has been lost due to the non-3D-consistent latent representation. This requires recovering discarded details rather than refining existing high-frequency information, making our optimization fundamentally more difficult than fixing artifacts in RGB space where pixel-level details are preserved. In settings where RGB-space Gaussian splatting produces satisfactory results, it may therefore be the preferable approach. 
Nonetheless, our method addresses critical scenarios where RGB rendering fails, as detailed in previous works~\cite{zhou2025latent, chen2024mvsplat360}, particularly in pipelines requiring latent space optimization for memory efficiency or compatibility with generative models. Additionally, our performance is inherently limited by the quality of the pre-trained VAE. Despite these limitations, our approach establishes a new paradigm for latent space 3D reconstruction, demonstrating that high-quality novel view synthesis is achievable through diffusion-based refinement in settings where latent rendering is necessary.

\section{Conclusion}
\label{sec:conclusion}
In this paper, we presented \paper, a novel approach for novel view synthesis from radiance fields, working in the VAE latent space of diffusion models. We analyzed the multi-view inconsistencies of this latent space, and identified several deficiencies in existing latent-space methods.
Our key insight is that these inconsistencies can be addressed by injecting high-frequency details from reference views through self-attention in a diffusion model.
Experimental results demonstrate that \paper consistently outperforms existing methods for latent novel view synthesis across all metrics. 
We further show that \paper can effectively enhance feed-forward latent 3DGS models, leading to more detailed and faithful renderings.

{
    \clearpage    
    \small
    \bibliographystyle{ieeenat_fullname}
    \bibliography{main}
}


\end{document}